# Deep Convolutional Neural Network-based Bernoulli Heatmap for Head Pose Estimation

Zhongxu Hu, Yang Xing, Chen Lv*, *Senior Member, IEEE,* Peng Hang and Jie Liu

*Abstract*—Head pose estimation is a crucial problem for many tasks, such as driver attention, fatigue detection, and human behaviour analysis. It is well known that neural networks are better at handling classification problems than regression problems. It is an extremely nonlinear process to let the network output the angle value directly for optimization learning, and the weight constraint of the loss function will be relatively weak. This paper proposes a novel Bernoulli heatmap for head pose estimation from a single RGB image. Our method can achieve the positioning of the head area while estimating the angles of the head. The Bernoulli heatmap makes it possible to construct fully convolutional neural networks without fully connected layers and provides a new idea for the output form of head pose estimation. A deep convolutional neural network (CNN) structure with multiscale representations is adopted to maintain high-resolution information and low-resolution information in parallel. This kind of structure can maintain rich, high-resolution representations. In addition, channelwise fusion is adopted to make the fusion weights learnable instead of simple addition with equal weights. As a result, the estimation is spatially more precise and potentially more accurate. The effectiveness of the proposed method is empirically demonstrated by comparing it with other state-of-the-art methods on public datasets.

## I. Introduction

Human-machine collaboration, such as autonomous vehicles and coexisting-cooperative-cognitive robots[3], is becoming a long-term trend and an active research topic[1,2,4,8]. Human understanding is the key to be solved. High accuracy and robustness of head pose estimation are crucial to many human-related tasks, including gaze detection, driver attention, human behaviour analysis, and fatigue detection. It can also be used to improve the performance for some face-related tasks, including expression detection and identity recognition.

Historically, there are several sensors available for head pose estimation, including RGB, depth, IR, IMU, and optical marker. The nonintrusive method based on vision is more easily accepted by users. The development of computer vision technology has also promoted an increasing number of researchers to adopt vision-based methods. Head pose estimation is a task that needs to infer the 3D information of the head from the input data. A naïve idea uses a depth sensor to obtain 3D input data [13,14]. Another idea uses spatiotemporal data such as video to supplement the temporal information lost in a single 2D image [27]. There are several geometry-based methods [5,6,16] that use a predefined 3D facial model to fit a single RGB or depth face image. The essence of geometry-based methods is to use the prior information of the head shape. As a registration optimization problem, the most critical task is how to obtain accurate feature matching. One idea is to obtain high-quality 2D facial features with the help of the facial landmark, and the other is to optimize or augment a predefined 3D model. Therefore, geometry-based methods are often a combination of multiple methods, and they require a predefined 3D model and combine some landmark-based methods. The landmark-based method means that it is often necessary to first obtain landmarks using the facial landmark detection model [7,17]. Some methods are inferred directly from these landmarks, and some are integrated into their own models. The problem is that the ground truth of the face landmark is work intensive. In this paper, we prefer to use a landmark-free method. What all these solutions have in common is the use of extra information. However, extra information means extra costs. Therefore, the goal of this paper is to adopt a low-cost solution, that is, to achieve head pose estimation based on a single 2D image.

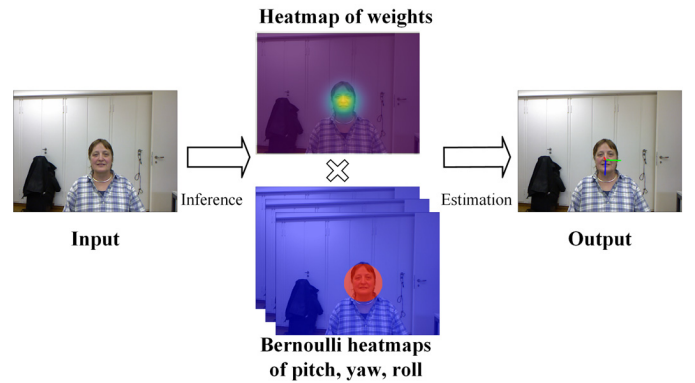

Fig. 1. The proposed Bernoulli heatmap. The pitch, yaw, roll correspond to different heatmaps.

The output of head pose estimation generally has two categories: direct regression[15] and converting to a classification problem, which can be called a "soft label problem"[3-4,9-12,28]. Allowing the network to output the

Z. Hu, Y. Xing, C. Lv and P. Hang are with the School of Mechanical and Aerospace Engineering, Nanyang Technological University, Singapore. (e-mail: {zhongxu.hu, xing.yang, lyuchen, peng.hang}@ntu.edu.sg)

J. Liu is with the School of Hydropower and Information Engineering, Huazhong University of Science and Technology, Wuhan, China. (e-mail: jie_liu@hust.edu.cn)

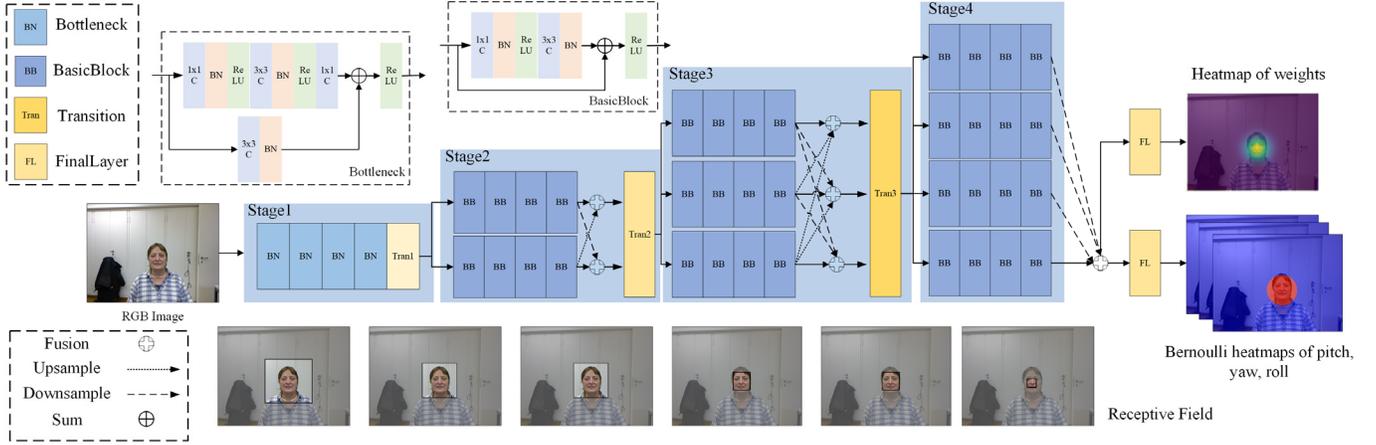

Fig. 2. The overall framework of the proposed method. The network is divided into four stages, where each stage has a different number of subnetworks. The subnetwork is composed of different numbers of **Bottleneck** or **BasicBlock** modules. These modules consist of several convolutional layers, batch normalization layers and activation layers. The **Transition** module is used for down-sampling and adding branches as the input for the next stage.

angle value directly for optimization learning is an extremely nonlinear process; the loss function's weight constraint will be relatively weak, and the spatial information of the feature map will be lost. When the output of head pose estimate is converted into a classification problem, the image is considered as a whole, so it is often necessary to preprocess the image first and crop out the head area; otherwise, the model is difficult to train. Human pose estimation also directly regresses the coordinates of joint points when CNN is first applied. However, heatmaps will soon be a commonly used method in human body pose estimation [18]. The basic approach is that one joint point corresponds to one heatmap. The advantage of this approach is that the output includes both classification and regression. The classification is divided into two levels: classifying the different heatmaps that distinguish the different joint points, and classifying the foreground and background in one heatmap. The regression of the position of the joint point actually belongs to a classification problem in a certain sense, that is, the pixel value on the heatmap indicates the probability that it belongs to a joint point, which can be regarded as a 2D version of one-hot encoding commonly used for classification tasks.

This paper is partially inspired by the Gaussian heatmap, but there are clear differences that make the Gaussian heatmap unavailable to our task. The output of the head pose is not the position of the joint point; it is the angle of the head. Therefore, the position or value of the maximum of the Gaussian heatmap cannot be used as the output of the head pose. This paper proposes a novel output, namely, the Bernoulli heatmap, which can achieve a fully convolutional network for head pose estimation, which focuses the network on the head area, as shown in Fig. 1. The value of the Bernoulli heatmap proposed does not conform to the Gaussian distribution but the Bernoulli distribution, and the final estimate is the average of nonzero values. It combines the two tasks of head detection and pose regression from some point of view, so it does not necessarily require preprocessing of the head crop.

As shown in Fig. 2, the network structure of multiscale representations is adopted, which is inspired by the HRNet [18]. The network has branches with different levels of resolution at different stages. The branches are composed of different numbers of bottleneck or BasicBlock modules. There is not only downsampling fusion through high resolution to low resolution but also upsampling fusion through low resolution to high resolution. In addition to the first stage, each stage includes feature extraction, multiscale fusion and stage transition. To better fuse the multiscale feature maps, channelwise fusion is used so that the fusion weights can be learned.

The main contribution of this paper is to propose a novel Bernoulli heatmap, which solves the problem that head pose estimation cannot construct a fully convolutional neural network. This will help to apply advanced models in many other fields more conveniently to head pose estimation and facilitate integration with human pose estimation to form an end-to-end whole-body pose estimation.

The remainder of this paper is structured as follows. Section 2 details our proposed method and network structure. Experimental settings and results analysis are presented in Section 3, followed by the conclusion and future work in Section 4.

## II. METHODOLOGY

In this section, the concept of the Bernoulli heatmap, the receptive field, the multiscale representations and the channelwise fusion are introduced.

### A. Bernoulli heatmap

Spatial generalization is an important capability in computer vision tasks, especially for the task of coordinate regression. Generally, the convolutional layers have spatial ability due to weight sharing, but the fully connected layers are prone to overfitting, thus hampering the spatial generalization ability of the overall network [20].

Gaussian heatmaps are currently the most common method used in the field of human pose estimation. The Gaussian heatmap can make the overall network into a fully convolutional network without the fully connected layer and ensure that a larger feature map can be output with a higher spatial generalization ability. Because the network is no longer required to convert spatial information into coordinate information by itself, it is easier to converge.

However, there are two problems with the Gaussian heatmap [19]. 1. There is a lower bound of theoretical error for the coordinate regression problem because its output is an integer. When the output map is reduced by n times the original image, there is a calculation error. When n is larger, the error is greater. 2. The Gaussian heat map uses the MSE loss function, which may cause an offset, as shown in Fig. 3(a). When the loss is low, it does not mean that the prediction is more accurate, but there may be deviations.

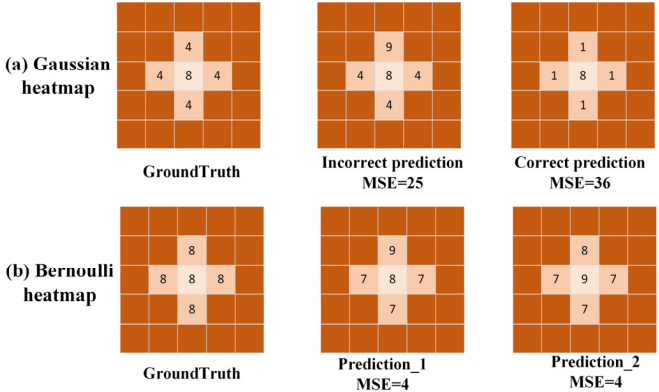

Fig. 3. The Gaussian heatmap(a)[19] and Bernoulli heatmap(b)

Head pose estimation not only regresses to the coordinates of joint points but also to three angles about head rotation. Therefore, the Gaussian heat map cannot be used directly. To solve this problem, the Bernoulli heatmap is proposed, and different angles correspond to different Bernoulli heatmaps. We define the ground truth of position p of the $ith$ Bernoulli heatmap, $L_i(p)$, as

$$L_i(p) = \begin{cases} v & \text{if } p \text{ in the } c \\ 0 & \text{otherwise} \end{cases} \quad (1)$$

where v is the value of the $ith$ angle, and c is a circle whose centre is the centre of the head with a radius $r$. $r$ is a hyperparameter. During testing, the angle $L_i$ is

$$L_i = \frac{1}{n(p)} \sum L_i(p) \quad (2)$$

where $n(p)$ is the mean number of nonzero $L_i(p)$.

The Bernoulli heatmap still focuses the model on the head area but does not care about the exact centre head point coordinates, so that there is no problem of theoretical error lower bound and deviations.

### B. Receptive field

The receptive field [19] is an important feature of the convolutional neural network. The value of each output node of the convolutional layer depends on a certain area of the input of the convolutional layer, as shown in Fig.4. Other input values outside this area do not affect the output value. For example, in classic object detection using an region proposal network (RPN), anchor is the basis of RPN, and receptive field (RF) is the basis of the anchor.

In [26], the theory of effective receptive field (ERF) was proposed. They found that not all pixels of the receptive field have the same contribution to the output vector or feature map. In many cases, the effective pixels of the receptive field conform to the Gaussian distribution, which only occupies a part of the theoretical receptive field, and the Gaussian distribution rapidly decays from the centre to the edge. This also means that the reliability of the predicted value of each pixel on the Bernoulli heatmap is different. The closer to the head region, the higher the reliability is, and the farther away from the head region, the lower the reliability is. Therefore, in addition to the Bernoulli heatmap, the final output of the model also has a Gaussian heatmap, which represents the probability of different positions away from the centre of the head and thus the weights of different positions, as shown in Fig. 1 and Fig. 2. The final estimation formula is as follows:

$$L_i = \frac{1}{\sum w_p} \sum w_p * L_i(p) \quad (3)$$

$$w_p \sim N(\mu, \sigma^2) \quad (4)$$

where $w_p$ represents the corresponding weight on the Gaussian heatmap, $\mu$ is the centre of the head area, and $\sigma$ affects the weight distribution of the heatmap. In this paper, $L_i$ is set equal to $0.6 * r$. The purpose is to make the effective area of the Gaussian heatmap smaller than the Bernoulli heatmap. In the experiments, only the pixels where $w_p$ is greater than 0.5 are used for the calculation.

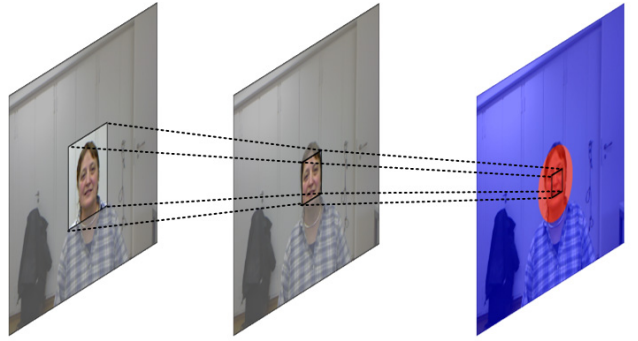

Fig. 4. The receptive field

In Section 3.1, the Bernoulli heatmap is discussed. The hyperparameter, the radius $r$, is relevant to the receptive field, especially to the ERF. It is not difficult to understand that when the head area is included in the receptive field, the angle will be predicted; otherwise, it is 0. This is equivalent to cropping the input image inside the network, including the head region, as positive samples, and the rest as negative samples.

### C. Network structure

#### 1) Multiscale representations

Many existing pose estimation networks are constructed by high-to-low resolution subnetworks in series, where each subnetwork is composed of a sequence of convolution layers and a down-sample layer to obtain low resolution. These methods need to recover the high-resolution from low-resolution representations.

Recently, the HRNet has better performance compared to others the HRNet is successful from two aspects: (i) maintaining the high-resolution representations through the whole network without recovering the high resolution from low resolution representations and (ii) fusing multiresolution representations repeatedly, rendering reliable high-resolution

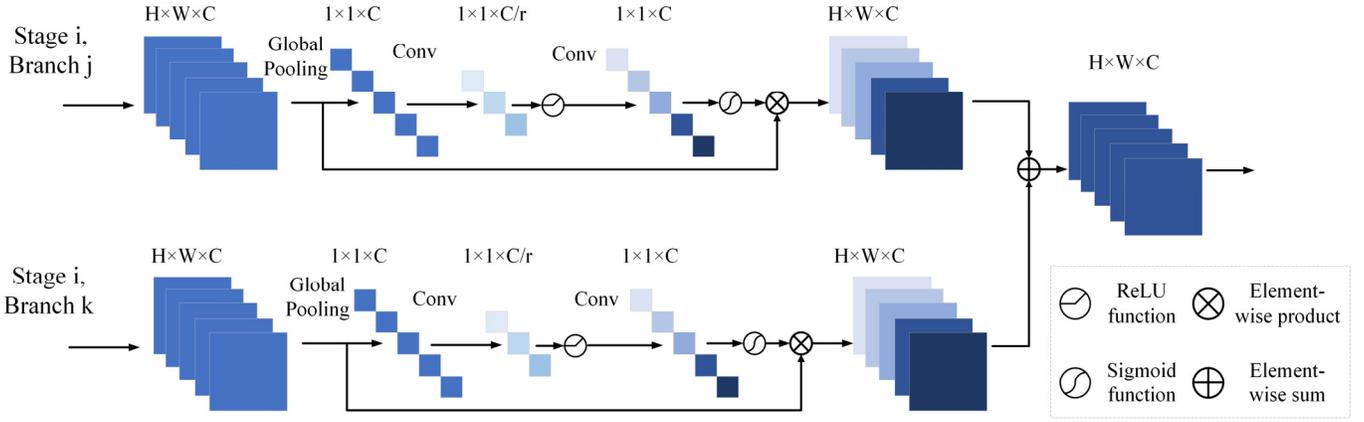

Fig. 5. The channel-wise fusion. Before the fusion of feature maps with different resolutions, the weight learning branch is added so that feature maps with different resolutions have different fusion weights.

representations.

In this study, we adopted this multiscale/multiresolution representation as our backbone. It is believed that the key capability of the convolution network is that the different level features can be learned, and these different levels of features have different contributions for the output vector. The current fusion is generally the addition of equal weights. However, we hope that their weights can also be learned during the fusion of feature maps.

*2) Channelwise fusion*

To make the network learn to fuse multiscale feature maps and exploit the interdependencies among feature channels, inspired by the research of Zhang et al. [22]. Channelwise fusion is used (see Fig. 5). Let $X = [x_1.., x_c.., x_C]$ be the feature maps of one branch before fusion, which has $C$ feature maps with the size of $H \times W$. Then, the $c$-th weight of $w$ is determined by

$$w_c = f(W_2 * \delta(W_1 * \frac{\sum_{i=1}^{H} \sum_{j=1}^{W} x_c(i,j)}{H \times W})) \quad (5)$$

where $x_c(i,j)$ is the value at position $(i,j)$ of c-th feature $x_c$. It can be seen that global average pooling (GAP) is first performed, which can be viewed as the aggregation of local descriptors to express the whole information. $W_1$ is the weight set of a convolution layer as channel-downscaling with reduction ratio $r$. Relatively, $W_2$ is the weight set of the channel-upscaling layer with ratio $r$, where $f(\cdot)$ and $\delta(\cdot)$ denote the sigmoid and rectified linear unit (ReLU) activation functions, respectively. The activation functions can obtain the channelwise dependencies from the aggregated vector, and then the nonlinear interactions between channels and the non-mutually exclusive relationship can be learned so that the multiple channelwise features can be emphasized instead of one-hot activation.

$$\widehat{x_c} = (1 + w_c) \cdot x_c \quad (4)$$

Then, we obtain the final channel weights $w_c$, which are used to rescale the feature map $x_c$. where $w_c$ and $x_c$ are the scaling factor and feature map in the c-th channel.

## III. EXPERIMENT

This section describes the experiment results, the dataset used, evaluation criterions, and comparisons with other methods.

### A. Dataset

Three different kinds of popular head pose datasets were used for the experiments: BIWI [21] and AFLW2000[33], whose sample is a real image with 3 pose angles, and 300W-LP [33], whose sample is a synthesized image with 3 pose angles. Some examples of these datasets are shown in Fig.6. Because the BIWI dataset is composed entirely of a large number of real images, this study used it as the main dataset for discussing the proposed method. However, considering diversity, AFLW2000 was used as a supplementary test set.

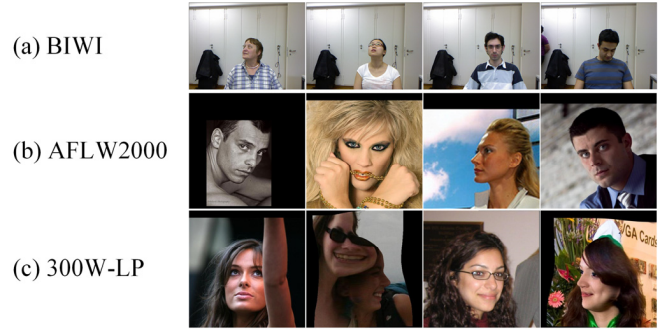

Fig. 6. The examples of the BIWI, AFLW2000 and 300W-LP datasets

**BIWI** The BIWI contains 24 sequences acquired with a Kinect sensor. Twenty people (some were recorded twice - 6 women and 14 men) were recorded while turning their heads, sitting in front of the sensor, at approximately one metre of distance. There are nearly 15,000 images in the database, and the angles include yaw: ± 75 degrees, pitch: ± 60 degrees, and roll: ± 50 degrees.

**300W-LP** The 300W across Large Poses (300W-LP) database contains 61,225 synthesized face samples from multiple alignment databases, including AFW, LFPW, HELEN, and IBUG, which was further expanded to 122,450 samples with flipping. For each original image, there are several synthesized pose faces with different degrees of yaw.

**AFLW2000** The AFLW2000 database contains the ground-truth 3D faces and the corresponding 68 landmarks of the first 2,000 AFLW samples. The samples in the dataset have large pose variations with various illumination conditions and expressions.

## B. Network Setting

As shown in Fig. 2, the network is divided into four stages, where each stage has a different number of subnetworks. The subnetwork is composed of different numbers of bottleneck or BasicBlock, and the number is shown in Tab. 1. The corresponding NUM_CHANNELS refers to the corresponding channel number of each bottleneck or BasicBlock module. The multistep learning rate decay was used, as shown in Tab. 1. INITIAL VALUE is the initial learning rate, LR_FACTOR is a multiplicative factor of learning rate decay, and LR_STEP is a list of epoch indices.

TABLE I
THE HYPERPARAMETERS OF THE NETWORKS

| Testing | BIWI | |
|---|---|---|
| | With Background | Without Background |
| Input Size | 128x96 | 64x64 |
| Stage1 | NUM_BOTTLENECK | NUM_CHANNELS |
| | 4 | 64 |
| Stage2 | NUM_BASICBLOCK | NUM_CHANNELS |
| | 4 | 32 |
| | 4 | 64 |
| Stage3 | NUM_BASICBLOCK | NUM_CHANNELS |
| | 4 | 32 |
| | 4 | 64 |
| | 4 | 128 |
| Stage3 | NUM_BASICBLOCK | NUM_CHANNELS |
| | 4 | 32 |
| | 4 | 64 |
| | 4 | 128 |
| | 4 | 256 |
| Learning Rate | INITIAL VALUE | 0.001 |
| | LR_FACTOR | 0.5 |
| | LR_STEP | 15 |
| | | 30 |

In addition to the sigmoid activation function used in channelwise fusion, the other activation functions were leaky ReLU, and the leaky value was 0.2. The optimizer was Adam during training.

## C. Results

In this study, two kinds of evaluation criteria were used to evaluate the estimation results. *Criterion I*: The mean absolute error (MAE) between the estimation and the ground truth of each angle in the test set, which gives a macroscopic assessment of the estimation result. *Criterion II*: The proportion of all three angles of all samples is less than a specific threshold. This criterion is more microscopic compared to the first criterion and is more concerned with local bad estimations.

To compare with others, we first used 16 videos of the BIWI dataset for training and the remaining 8 videos for testing. In Section 3, we mentioned the hyperparameter $r$. To study its impact on the results, we performed several sets of experiments, as shown in Fig. 7. AT denotes channelwise fusion. BG denotes the situation with background; in contrast, nBG denotes the situation without background. The first group of experiments is that the network has channelwise fusion with different $r$, and the other group of experiments is that the network does not have channelwise fusion. It can be seen that $r$ has an effect on the results. When $r$ was too large, from the perspective of the receptive field, the effective receptive field of some effective pixels on the heatmap did not contain valid head information, which means that the "negative sample" was regarded as a "positive sample", which inevitably led to large errors. Similarly, when $r$ was too small, some "positive samples" were regarded as "negative samples" at this time, which also inevitably led to large errors.

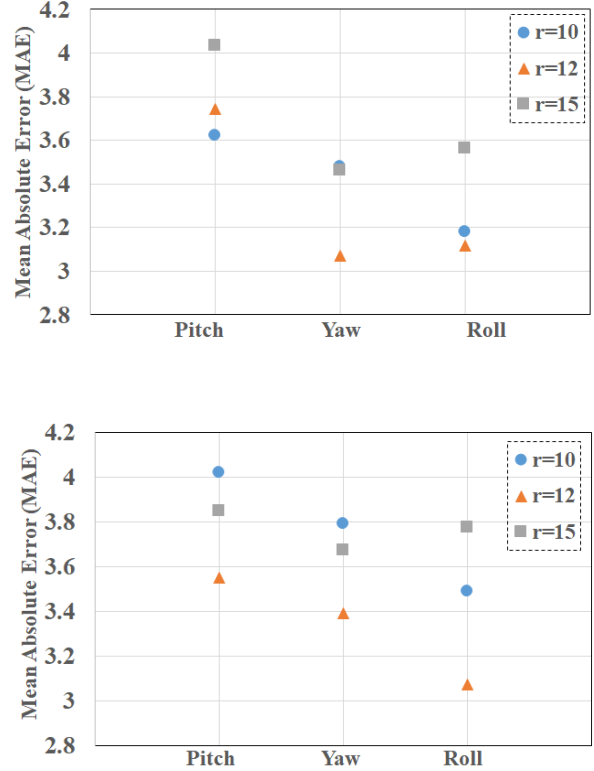

Fig. 7. The comparisons of different hyperparameter r in case of *Criterion I*: HR-AT-nBG (up) and HR-nBG (Down)

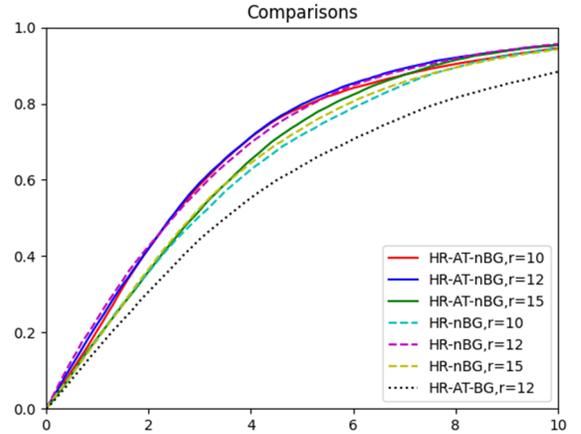

Fig. 8. The comparisons of different hyperparameter $r$ in case of *Criterion II*. The horizontal axis represents the error threshold (degree), and the vertical axis represents the percentage under the corresponding threshold.

As shown in Fig. 7 and Fig. 8, when the basic parameters are the same, channelwise fusion is beneficial to the model. In Fig. 7, our methods are basically convex curves. It can be roughly seen that when the threshold is less than 3, the slope of the curve is large, and the percentage at this time is greater than 50%, after which the slope gradually decreases. When the threshold is approximately 5, the percentage is approximately 80%. This means that the overall error of our methods is lower, but the presence of a small number of samples with larger errors makes the average error higher.

In addition, to test the robustness of our method, the translation and occlusion of the test set were used without retraining the model, as shown in Fig. 9. Each case was tested five times, and each test sample was randomly processed each time. The test results are shown in Fig. 10. When there was some occlusion and translation, although the MAE has increased, the increase was not large. We believe that if the training samples were augmented and then trained, the model could handle more severe translation and occlusion problems.

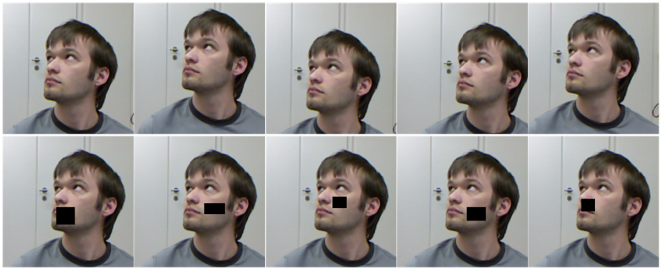

Fig. 9. Some examples of the translation and the occlusion

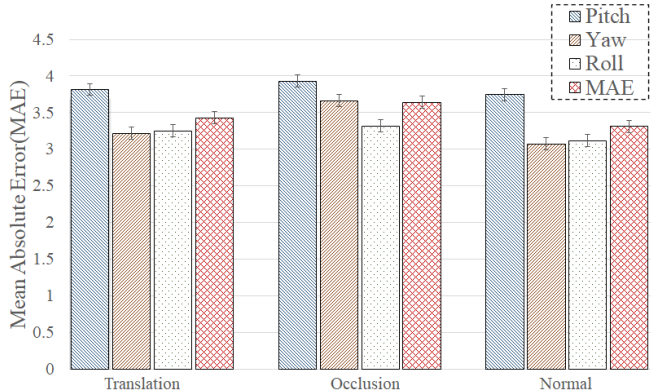

Fig. 10. The results after different processing of test set.

For neural networks, the most important feature is generalization. To evaluate the generalization of the proposed method, we used three protocols. Protocol 1: As before, 16 videos (~10K) of the BIWI dataset were used for training, and the remaining 8 videos (~5K) were used for testing. Protocol 2: Approximately 12 videos (~7.5 K) of the BIWI dataset were used for training and the remaining 12 videos (~7.5K) for testing. Protocol 3: Approximately 8 videos (~5K) of the BIWI dataset were used for training, and the remaining 16 videos (~10K) were used for testing. As general knowledge, the error was lower when there were more training samples. However, it can be seen from Fig. 11 that even when there were few training samples, the test results remained in a relatively reasonable range. This shows that the proposed method has good generalization. When the training sample increased, better results were obtained.

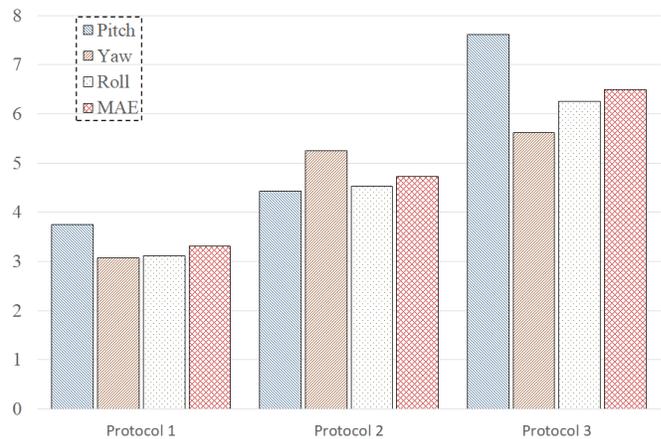

Fig. 11. The results of different protocols

### D. Comparisons

For the BIWI dataset, we used 16 videos of the BIWI dataset for training and the remaining 8 videos for testing, similar to others. As shown in Tab. 2, DeepHeadPose[23] focused on lower solution multimodal RGB-D images, which combined classification and regression to estimate approximate regression confidence. Gu et al. [25] used modified VGG16 and FC-RNN based on Bayesian filters to estimate the 3 angles, and they used the RNN to handle the sequence images. The SSR-Net [24] took a coarse-to-fine structure, where each stage performed multiclass classification. FSA-Net [12] was proposed to learn a fine-grained structure mapping for spatially grouping features before aggregation, which provided part-based information and pooled values. Martin et al. [25] used depth information in addition to RGB information. These methods crop the background and keep the face area. Therefore, there are two situations for our approach: one is to remove the background, and the other is to retain the background.

TABLE II
THE COMPARISONS WITH THE-STATE-OF-ART METHODS

| Dataset | BIWI | | | |
|---|---|---|---|---|
| Method | Pitch | Yaw | Roll | MAE |
| DeepHeadPose[23] | 5.18 | 5.67 | - | - |
| SSR-Net-MD[24] | 4.35 | 4.24 | 4.19 | 4.26 |
| VGG16[25] | 4.03 | 3.91 | 3.03 | 3.66 |
| VGG16+RNN[25]* | 3.14 | 3.48 | 2.6 | 3.07 |
| Martin [29]† | 2.5 | 2.6 | 3.6 | 2.9 |
| FSA-Caps-Fusion[12] | 4.29 | 2.89 | 3.6 | 3.6 |
| **HR-AT-BG, Ours** | 5.25 | 4.91 | 4.25 | 4.8 |
| **HR-nBG, Ours** | 3.55 | 3.4 | 3.07 | **3.34** |
| **HR-AT-nBG, Ours** | 3.74 | 3.07 | 3.11 | **3.31** |

'–' denotes the corresponding results are unavailable in their papers. * The method uses time information. † The method uses depth information.

As shown in Tab. 2, when the input image was a single RGB image, that is, when depth information and time information were not considered, and the background was cropped similar to the other methods, our method performed better. When the background was retained, our method still maintained a low error.

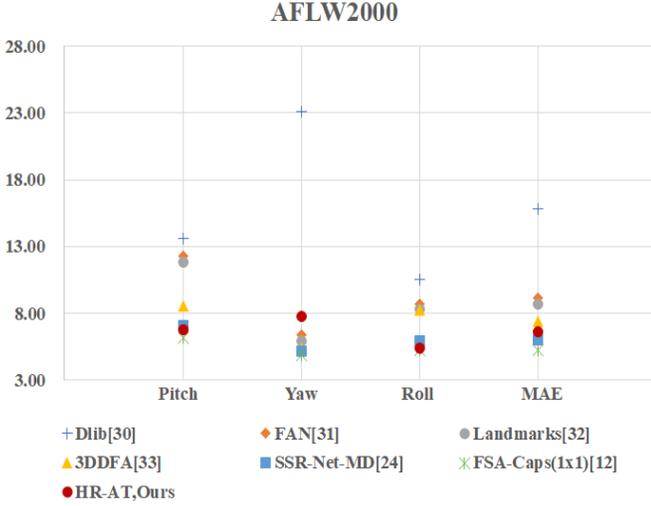

Fig. 12. The comparisons with others in the AFLW2000 dataset

For further comparison, the synthetic dataset 300W-LP was used as the training set, and AFLW2000 was used as the test set. The mean and standard deviation of ImageNet were used to normalize the colour channels of the samples. The remaining network hyperparameter settings were basically the same. Compared with several state-of-the-art methods, the results are shown in Fig. 12. **Dlib** [20] is a face-related library that contains face and landmark detection and several other functions. It can be used as a benchmark for landmark-based methods. The **FAN** [31] is a state-of-the-art landmark-based method. It is robust against occlusions and different poses. The multiscale information is used by merging multiple features across different layers. **Landmarks**[32] retrieves head poses from the ground-truth landmarks of the AFLW2000 dataset. **3DDFA** [33] is a geometry-based method that fits a 3D standard head model to an RGB image and allows robust alignment of the landmarks of the head. **SSR-Net-MD**[24] and **FSA-Caps(1x1)**[12] use a hierarchical soft stagewise classification and attention module for pose estimation. In terms of the MAE of the AFLW2000 benchmark, only these two methods are better than ours, and the gap is not large. These two methods also use the concept of soft labels.

To further analyse the performance of the proposed method on the AFLW2000 dataset. A new evaluation criterion was introduced. *Criterion III*: The MAE of the absolute value of the pose angles of the sample within a certain range. This evaluation criterion measures the performance of the method in different angle intervals. Based on this evaluation criterion, the proposed method was compared with the **SSR-Net-MD** and **FSA-Caps(1x1)**. It can be seen in Fig. 13 that when the angle range of the sample was small, the MAE of our proposed method was significantly lower, and the proportion of such samples was larger, reaching almost 90%. This means that a few samples with larger angles lead to an increase in the total MAE of our proposed method. Therefore, we believe that if more such samples are collected, the MAE of the proposed method can be reduced.

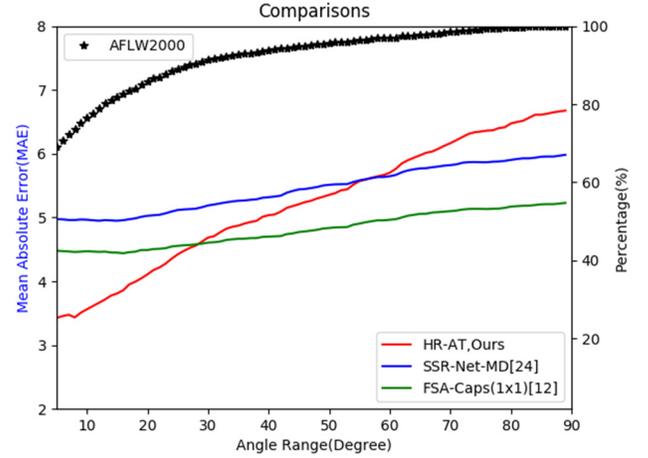

Fig. 13. The comparisons with two state-of-the-art methods with *Criterion III*. The black asterisk curve of AFLW2000 corresponds to the vertical coordinate on the right which represents the percentage of samples in different angle ranges, the remaining three solid curves correspond to the vertical coordinates on the left which represents the MAE in different angle ranges, and the horizontal coordinate is shared.

*E. Visualization*

To further understand the proposed method and influence of the Bernoulli heatmap on intermediate feature maps, the feature maps were visualized, as shown in Fig. 14. The feature maps after each stage were normalized and merged. It can be seen that the activation of the head region was stronger, and its range increased as the stage increased. Our visualization method adds feature maps of different resolutions with equal weights. This does not affect the overall understanding of our approach.

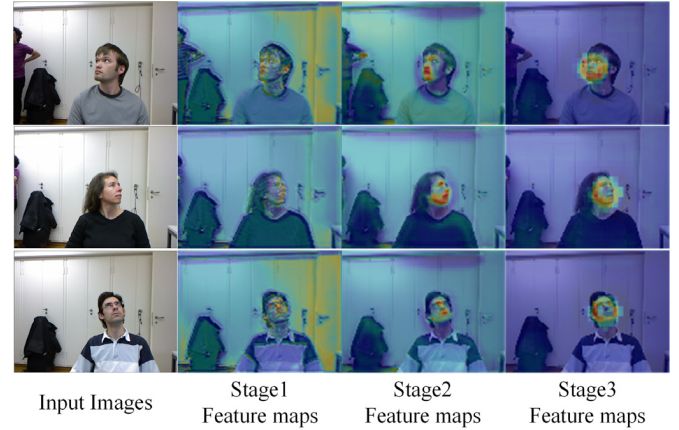

Fig. 14. The visualization of the intermediate feature maps

We randomly selected several samples and visualized the estimated results in the BIWI dataset, as shown in Fig. 15. It can be found that regardless of whether the background was cropped, compared with the ground truth, although there was an error in the value, there was no obvious difference from an intuitive point of view. Compared with other methods that require face detection, our method could directly locate the head and give the corresponding angles of the head. In addition, some AFLW2000 images were randomly selected for visualization. These samples had large pose variations with various illumination conditions and expressions. As shown in Fig. 16, the examples included the estimation and the ground truth. In most of the samples with small angles, the estimation and ground truth basically coincided. There were deviations in a few samples with large angles, but they were within a reasonable range.

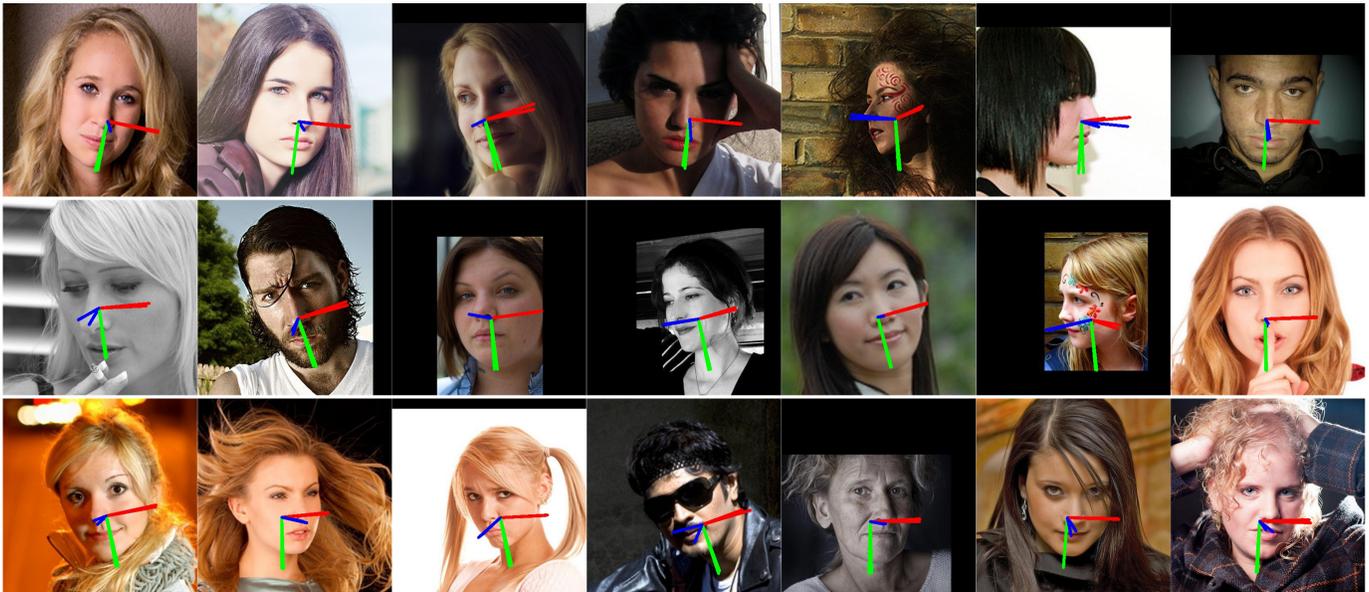

Fig. 16. The estimation examples of the AFLW2000 dataset. These examples include the estimation and ground truth.

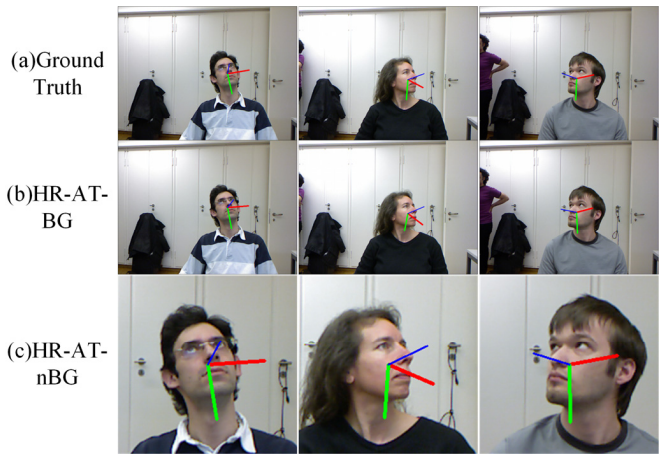

Fig. 15. The examples of our result in the BIWI dataset

To understand why the proposed method is not good for large-angle samples, some large-angle samples were selected from the 300W-LP and AFLW2000 datasets. As shown in Fig. 17, because the 300W-LP is a synthetic dataset, the samples with large angles have severe distortions and thus have a large difference from the real images of the AFLW2000 dataset. The features of the relevant area are distorted and blurred. The difference between the training set and the test set on the large-angle samples led to poor performance of the proposed method in this part of the sample. Another reason is that when the head angle was large, that is, when turning to the side, two people often appear at this time, as shown in the second and fifth columns of Fig. 17(b), which will affect the inference of the model and cause the error to increase. Through these visualization methods, a further understanding of our proposed method will also help to find subsequent optimization and research directions.

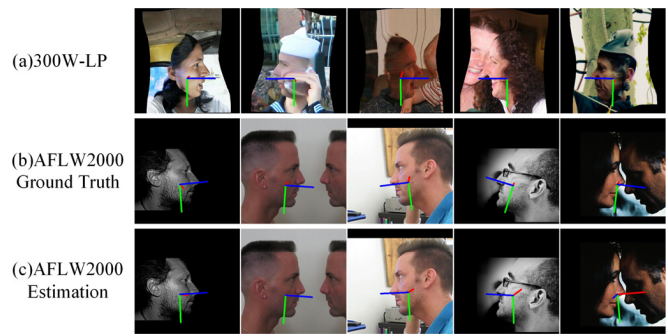

Fig. 17. The large angle examples of 300W-LP and AFLW2000

## IV. Conclusion

In this paper, we proposed a new output called the Bernoulli heatmap for the head pose estimation task, which makes the network a fully convolutional network and provides a new idea for head pose estimation. The Bernoulli heatmap not only regresses the angles of the head but also allows the network to distinguish between the foreground and the background, thereby improving the robustness to the background. This benefits integration with human pose estimation to form an end-to-end whole-body pose estimation. We adopted a multiscale representation network structure that is similar to HRNet to maintain high-resolution representations. The difference is that channelwise fusion was used so that the fusion weights of feature maps with different resolutions could be learned. The experiments show that our methods performed well compared to the state-of-the-art methods in public datasets.

In future work, we will improve the performance of the proposed method on large-angle samples. We think that our method can still continue to be optimized and that the Bernoulli heatmap can be used for other regression tasks.

ACKNOWLEDGMENT

This work was supported by the A*STAR Grant (No. 1922500046), Singapore.